# Enhancing LLM with Evolutionary Fine-Tuning for News Summary Generation

Le Xiao, Xiaolin Chen

College of Information Science and Engineering, Henan University of Technology, Zhengzhou, China

xiaole@haut.edu.cn  chenxiaolin@stu.haut.edu.cn

**Abstract**

News summary generation is an important task in the field of intelligence analysis, which can provide accurate and comprehensive information to help people better understand and respond to complex real-world events. However, traditional news summary generation methods face some challenges, which are limited by the model itself and the amount of training data, as well as the influence of text noise, making it difficult to generate reliable information accurately. In this paper, we propose a new paradigm for news summary generation using LLM with powerful natural language understanding and generative capabilities. We use LLM to extract multiple structured event patterns from the events contained in news paragraphs, evolve the event pattern population with genetic algorithm, and select the most adaptive event pattern to input into the LLM to generate news summaries. A **N**ews **S**ummary **G**enerator **(NSG)** is designed to select and evolve the event pattern populations and generate news summaries. The experimental results show that the news summary generator is able to generate accurate and reliable news summaries with some generalization ability.

## 1. Introduction

In recent years, pre-trained language models have undergone rapid development [1, 2, 3, 4, 5, 6], especially models based on the Transformer [7] architecture, which has the ability to process natural language. These models are able to automatically learn statistical patterns and patterns in language by training on large-scale textual data, which makes pre-trained language models widely adaptable and can be applied to a variety of domains and tasks.

Large language models (LLM) have improved the experimental results of many natural language processing tasks, exceeding the previous state-of-the-art of deep learning models in tasks such as information extraction [8] and causal inference [9], and therefore, how to enhance the performance of LLMs in specific tasks has attracted extensive research. News summary generation is a type of document summary generation [10], which aims to generate concise and important event topics in a paragraph of text to better communicate intelligent content. It plays a key role in areas such as information processing, intelligence analysis, research, and decision-making. Automatic news summary generation can provide accurate and comprehensive information to help people better understand and respond to complex real-world events.

Traditional news summary generation is mainly used to generate headlines, a task that requires the model to be able to understand the key information of the news text and be able to express it in a concise manner.

Previous news headline generation methods [11, 12, 13] currently have some challenges, due to the limitations of the model itself and the amount of training data, it is often difficult for the model to fully understand the semantics of the article, and if there is too much noise in the representation of events and the quality of the data set is low, it may affect the consistency of the generated headlines with the original text, and it is often difficult to accurately generate reliable headlines for articles in certain specialized fields, event pattern as a kind of structured data with concise and accurate characteristics. To address the above challenges, we use LLM to generate news summaries. News paragraphs often have long text segments and contain multiple events, each with an implied event pattern, as in Figure 1, to enhance the summary generation capability of LLM through event pattern evolution.

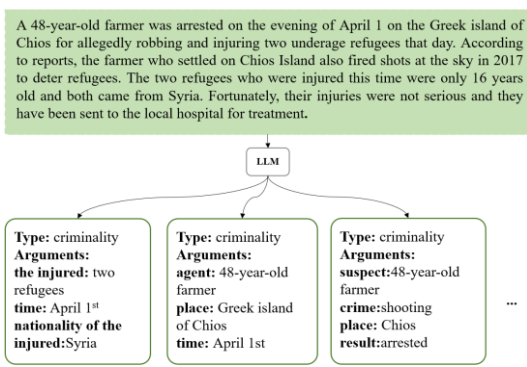

Figure 1: Get event patterns from news paragraphs

We propose a new paradigm for news summary generation - using LLM to obtain event patterns from texts to construct a pool of event patterns, and with the help of the idea of genetic algorithm, we use the event patterns as chromosomes, select crossover of chromosomes to evolve the pool of event patterns, and evaluate the chromosomes in the pool by the fitness function The highest fitness means the best quality, and the best quality individuals are selected to enhance LLM so that LLM generates accurate, reliable and comprehensive news summaries.

To this end, we designed a news summary generator (NSG) to automatically generate accurate and comprehensive news summaries in a three-step process:

1. Unsupervised candidate event pattern pool generation. In the unsupervised case, multiple events are extracted from the news text and a pool of event patterns is constructed based on contextual fine-tuning of the LLM.

2. Candidate event pattern population evolution. A genetic algorithm is used to evolve the event pattern pool by selecting crossover for event patterns.

3. Fine-tuning enhanced LLM to generate news summaries. The most adapted event pattern is selected to generate news summaries using LLM.

Main contributions to this article:

We propose a new news summary generation paradigm, i.e., with evolutionary knowledge, fine-tuning LLM for enhancement to improve the news summary generation capability of LLM and effectively address the challenges of inaccurate, incomplete, and poorly interpretable news summaries generated by LLM.

Design a news summary generator to generate candidate event pattern pools from open-world knowledge using LLM, iteratively update the event pattern individuals to evolve the candidate event pattern pools, and select the most adaptive individuals to enhance LLM to generate accurate and comprehensive news summaries.

Experiments have proven that news summary generators can generate accurate and comprehensive news summaries from news texts, providing concise and accurate natural language descriptions to better understand and convey the core content of events, and are valuable for intelligence analysis.

## 2. Background and Related Work

This section reviews related research on news summary generation, event extraction, and knowledge acquisition using LLM. Genetic algorithms are also introduced.

### 2.1 Research on Event Extraction

Event extraction is an important task in the field of natural language processing, aiming at extracting key information about events from text. The event extraction task can be divided into two main areas:

1. Trigger Identification: The goal of this task is to identify keywords or phrases in the text that indicate the occurrence of an event, also known as event trigger words. Trigger words are usually verbs or noun phrases that can explicitly or implicitly indicate the occurrence of an event. Identifying trigger words is the first step in event extraction and helps to identify the events present in the text [14].

2. Event Argument Identification and Classification: Once the event trigger word is identified, the next task is to identify the arguments associated with the event, i.e., information about the participants, time, and place of the event. The event argument elements can be noun phrases or verb phrases, which have relationships with the event trigger words, such as subject-action, action-object, etc.

The event extraction task's challenge is dealing with semantic complexity, polysemy, and context dependency. A combination of techniques such as lexical annotation, dependent syntactic analysis, entity recognition, and semantic role annotation is required to identify event trigger words and event thesis elements and to classify and associate them. Event extraction is of great value in applications such as information extraction, text understanding, and knowledge graph construction, and helps to extract and organize event information from large-scale text data to support various practical application scenarios.

Traditional machine learning-based event extraction tasks can be divided into pipeline-based event extraction and union-based event extraction according to the solution process of these above tasks [15]. The pipeline-based approach treats all subtasks as independent classification problems: [16, 17], and the joint-based approach [18, 19].

The set of event types and their meta-roles obtained from event extraction constitutes an event model that helps to completely describe the event and understand its aspects, in the form of "type: bombing; meta-roles: perpetrator, victim, target, tool" [20].

The event schema consists of two parts: event type describes the general category or set of categories to which the event belongs, and meta-roles refer to the semantic roles or functions associated with the event in different parts of the event schema. Each theoretical meta-role represents an element or role in the event, such as the performer of the action, the bearer of the action, the object of influence of the action, etc. They describe the relationships and roles between different participants in the event. Argumentative meta-roles are usually abstract and aim to represent universal concepts or role types in an event. Common meta-roles include Subject, Object, Time, Place, Actions, etc. By defining and identifying different theoretical meta-roles, the components of an event can be more accurately described and the semantic analysis and understanding of the event can be performed.

Event patterns provide a structured way to represent and describe events. Through event patterns, we can capture important information about events, understand and analyze the core content, relationships, and characteristics of events, and identify the key elements and players of events.

Natural language generation is based on event patterns, i.e., generating corresponding natural language descriptions or sentences based on event patterns to generate specific and accurate representations of events for generating news headlines, summaries, reports, etc.

## 2.2 Genetic Algorithm

Genetic algorithm [22] is an optimization algorithm based on the theory of biological evolution, which simulates the evolutionary process in nature and searches and optimizes the solution space of a problem through genetic operations (e.g., selection, crossover, and mutation).

The basic idea is to create a set of individuals (called a population), each of which represents a candidate solution to the problem. By iterating generation by generation, the solutions in the population are improved by selecting good individuals, crossover and variation operations, and gradually approximating the optimal solution.

In this paper, we use LLM to generate original populations and evolve individuals by selection and crossover in the following process:

1. Initial population generation: LLM generates event patterns as individuals from event representations based on contextual fine-tuning unsupervised, and constructs a pool of candidate event patterns as the original population.

2. Adaptation assessment: Each individual is assessed using an adaptation function that measures accuracy and comprehensiveness (metrics).

3. Selection: Based on the evaluation results of the fitness function, the individual with higher fitness is selected as the parent solution and used to generate the next-generation solution.

4. Crossover: The gene crossover operation is performed on the newly generated solutions to increase the exploitability of the solution space by introducing randomness.

5. Update the population: replace or merge the newly generated individuals with the parent population to form an updated pool of event patterns.

6. Repeat steps 2 to 5: iteratively perform selection and crossover operations until the termination conditions are met.

After the evolutionary process of the genetic algorithm, the event patterns with the highest fitness are filtered out in the pool of candidate event patterns. These event patterns have high fitness values and quality to summarize the text's important events and key information.

## 2.3 Text Summary Generation

Natural language expressions of events are diverse and sparse, and summarizing news summaries, i.e., unifying different event representations and presenting the core content of events with concise and accurate natural language descriptions, is of great value for intelligence analysis. Summarizing text can help one quickly understand and organize large amounts of intelligence data, extract key information, avoid drowning in noise, and thus understand and convey the main points of an event more effectively. This unified and summarized description helps eliminate information redundancy and ambiguity, improves the efficiency and accuracy of intelligence analysis, and provides strong support for decision-making.

By summarizing news summaries, a large amount of heterogeneous information can be better processed and utilized to provide powerful guidance and decision support for intelligence work. Often, intelligence analysis within niche domains has similar needs, requiring the abstraction of a unified summary from a large number of similar events.

Traditional news summaries are generated using rules and templates designed by human experts to generate news summaries and probability-based methods, including the use of text clustering to group related events into pre-defined summaries based on manually edited summaries, in addition to methods that use the longest common subsequence to generate a representation of news summaries [11] but are more affected by data sparsity.

With the development of deep learning techniques, neural network-based approaches [23] have gradually become the mainstream of news summary generation, which can better handle the semantic and contextual complexity and overcome the problem of data sparsity, but often lead to overfitting due to the unavailability of large-scale training data [24].

With the rise of LLM, they have shown good results on many natural language processing tasks, but the black-box nature of LLM causes a lack of interpretability in the results they generate, which makes it difficult to gain insight into how the models generate news summaries from the input text data.

The interpretation of event patterns can also provide a basis for evaluating and interpreting the generated news summaries. Using event patterns to explain the generation of news summaries can help us understand how LLM generates high-quality news summaries based on event-related information, and we can gain a clearer understanding of how the models understand and represent key features, semantic relationships, and contextual information about events.

**2.4 Knowledge Acquisition with LLM**

The advantage of using LLM for knowledge acquisition lies in their ability to understand and generate language and to learn and extract knowledge from large-scale text data.

Cao [21] et al. proposed five cycles of knowledge in LLM: acquisition, representation, exploration, editing, and application.

In the field of news summary generation, LLM can generate accurate and concise text summaries by understanding and semantic analysis of text. It can help extract the key information and core content in the text, thus enabling knowledge distillation and acquisition. However, it should be noted that LLM also has some limitations, such as the quality of the generated news summary representations, and challenges such as accuracy and coverage.

To improve the quality of LLM-generated news summary representations and to address the challenges of accuracy and comprehensiveness, this paper proposes an LLM-based news summary generation paradigm and designs a news summary generator based on it to generate high-quality, accurate, and comprehensive news summary representations from a large number of event representations.

**3. News Summary Generation**

This section describes the method of news summary generation using the large language model.

Since a piece of news text often has a large amount of data, direct generation of event patterns may lead to the introduction of noise and phantom data due to too many argumentative roles and may cause important argumentative elements to be missing, so we extract different patterns from the different event contents contained in it and use a genetic algorithm to evolve the extracted event patterns and evaluate the final patterns by the fitness function, and the pattern with the highest score The highest scoring patterns are input to LLM to generate news summaries.

Formally, given the corpus and LLM, there are N news fragments in the corpus { $T_1$, $T_2$, ..., $T_N$ }, each news fragment can extract multiple events, LLM obtains an event pattern from each event in the news fragment $T_n$, n ∈ N and constructs a pool of event patterns $p_n$ as the original population for evolution, $p_n$ ={$s_n^1$, $s_n^2$, ... ,$s_n^i$}. $s_n^i$ represents the i-th event pattern in the n-th event pattern pool.

In this paper, we use a genetic algorithm to evolve the population and select the crossover operations of the theoretical roles contained in the individuals. Based on the above method, we design the News Summary Generator (NSG) to enhance the LLM news summary generation, and the framework is shown in Figure 2.

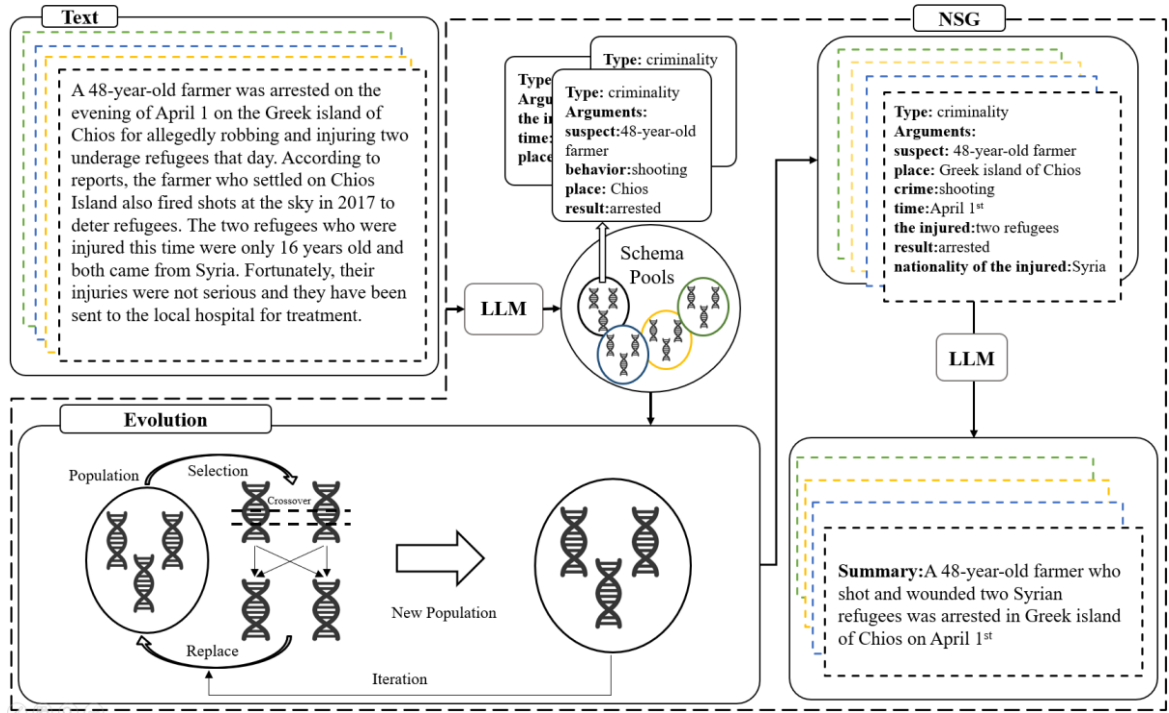

Figure 2 The Framework of NSG

The News Digest Generator contains three components:

1. Candidate event pattern pool generation.
2. Event pattern population evolution.
3. News summary generation.

In the following, we will describe these sections in detail.

### 3.1 Candidate Event Pattern Pool Generation

Event patterns are the basis for generating news summaries, providing abstraction and generalization of event representations. By acquiring event patterns, key information, themes, and core contents can be extracted from a large number of event representations, thus enabling efficient understanding and summarization of events.

Real-world text data usually contains noise and diversity, including misspellings, irregular grammar, randomness, etc. These factors increase the difficulty of accurately extracting event patterns from text, requiring models with certain robustness and generalization capabilities. Event patterns often exist in the implicit information of text rather than explicit representations, and the model needs to be able to understand contextual dependencies and infer the implicit event information in order to capture event patterns accurately. Traditional neural network-based event pattern extraction methods to acquire event patterns require a large amount of training data, and in some domains or specific event types, relevant data may be very limited, and this data scarcity can limit the model's ability to learn and understand event patterns.

LLM is equipped with powerful language generation and context learning capabilities through the training of large-scale text data. Based on this, this paper uses LLM to unsupervised and automatically generate candidate event patterns from the original event representations.

Based on this framework, we use LLM for contextual fine-tuning. First let LLM select an overall event type for the news to generalize, then identify different events from the text and get as many event patterns as we need based on each event. The purpose of context fine-tuning is to improve the performance and generalization of the model on a specific task and to generate results as we need them. The advantage of context fine-tuning is that it can take advantage of the general language knowledge already learned by the pre-trained language model and fine-tune it on a specific task, and by fine-tuning the model on task-relevant data, it allows the model to better understand the task-specific semantics and context and to generate task-relevant output.

Each context is a text pattern pair containing an event representation and the event patterns obtained from that event representation, which is used to guide the LLM for a generation. LLM unsupervised acquires event patterns from different types of events based on contextual fine-tuning, and constructs a pool of N event patterns based on N news segments in the corpus, with each event pattern pool evolving as a primitive population.

**3.2 Knowledge Evolution**

In this paper, we use the idea of genetic algorithm to generate news summaries by combining good genomes between individuals of different parents through genetic recombination to produce offspring with higher adaptability, select crossover by using the theoretical meta-role in event patterns as genes, and finally select the event pattern with the highest adaptability from the population. The pool of event patterns generated in the previous section provides the raw material for the evolution of event pattern selection.

In this paper, a pool of N event patterns obtained from N news clips is used as the original population, and the event patterns in the event pattern pool $p_n$ corresponding to news clips $T_n$, $n \in N$ are used as chromosomes, which are formally the set of event types and argument roles: {Type; Arguments}. In each generation, all chromosomes in the population need to be evaluated using the fitness function. Chromosomes with higher fitness values are placed in the mating pool, replacing chromosomes with lower fitness to update the population.

In this paper, we evaluate the salience and reliability of the thesis meta-role in terms of frequency of occurrence and importance, respectively, and design the fitness function. The salience and reliability of the thesis meta-role determine the accuracy and comprehensiveness of the generated summary.

The TF-IDF score is the event pattern $s_n^i$ ratio of the frequency of occurrence in the current population to the frequency of occurrence in the full population. The formula is as follows:

$$F(s_n^i) = (1 + \log(freq(s_n^i))^2) * \log\left(\frac{N}{\sum_n^N freq(s_n^i)}\right) \quad (1)$$

$freq(s_n^i)$ for the thesis meta-role $s_n^i$ frequency of occurrence in $P_n$.

The TextRank [25] score treats a document as a network of words and the links in the network as semantic relationships between the theoretical roles, with the following equation:

$$W(s_n^i) = (1-d) + d * \sum s_n^j \in I_n(s_n^i) \frac{w_{ij}}{\sum s_n^k \in Out(s_n^j) w_{jk}} * W(s_n^j) \quad (2)$$

The fitness function Q is defined as the weighting of the TFIDF score and TextRank score of the argumentative meta-role in the event pattern. The formula is as follows:

$$Q = \alpha * F(s_n^i) + \beta * W(s_n^i) \quad (3)$$

α and β is are the hyperparameters, and d is the damping factor.

Chromosomes with the highest scores are considered to be the best current solution. The selection of chromosomes with higher fitness allows the genetic algorithm to select the highest-quality event patterns.

For the selection of parents, the fitness of each event pattern in the population was first evaluated according to Equation (3), and a portion of individuals were selected as parents to participate in the next evolution according to the fitness value using the roulette algorithm [26], where event patterns with higher fitness have a higher probability of being selected, while event patterns with lower fitness still have a certain probability of being selected, which is conducive to maintaining the diversity of the population.

Since the event patterns contain fewer and less diverse argument roles, introducing variation operations at this time may introduce unnecessary randomness, so using only crossover operations here can maintain the stability of the population and allow faster convergence to the optimal solution. The crossover operation combines the chromosomes in the pool randomly two by two into event pattern pairs and swaps its own k argument roles, k being a random number no greater than the minimum number of argument roles contained in the two event patterns.

Subsequently, the progeny individuals obtained by crossover are merged with the parent individuals to form a new generation of population. The new generation population is involved in the selection and crossover operation as the parent of the next iteration. In this paper, we set the maximum number of iterations I, I am the hyperparameter, and the algorithm stops running when this number is reached, and finally, the event patterns in the population are ranked according to the fitness value, and the individual with the highest fitness value is selected to represent the most accurate and comprehensive event pattern with the highest quality as a hint for news summary generation.

### 3.3 News Summary Generation

We use 3.2 to obtain highly adaptable event patterns and input the highly adaptable event patterns into the LLM to generate more accurate and comprehensive news summaries. The black-box characteristics of the LLM also cause the generated results to often lack interpretability. The event model can provide explanations for the news summaries generated by the LLM, thus increasing their interpretability and making the generated results easier to understand and interpret.

In the second step, we extract a large number of event patterns from the event representations contained in each news clip and evolve the population continuously, and after a certain number of evolutions sort the event pattern with the highest fitness s as the high-quality event pattern. n news clips are obtained with a total of N high-quality event patterns $S = \{s_1, s_2, ..., s_N\}$. The set of high-quality event patterns is then input into the large language model to obtain news summaries.

The quality of the generated news summaries is assessed by a series of metrics such as ROUGE, BLEU, and Overlap scores.

By combining high-quality event patterns with the generative power of large language models, we are able to obtain more accurate and comprehensible news summaries, resulting in better results in terms of information conveyance and understanding. This approach not only improves the quality of news summaries but also enhances our understanding of the news summary generation process.

## 4. Experiment

### 4.1 Experimental Preparation
Dataset:
We experimented with the PENS [27] news headline generation dataset, which contains 113,762 news articles in 15 topic types, each containing id, headline, content and a category tag, of which we only used the news headline and content. The grain storage pest event corpus constructed by Xiao Le [28] et al. is also used as a niche domain dataset for comparison to explore the reliability of the summaries generated by the news summary generator proposed in this paper on a niche specialized domain, which is a grain storage pest event corpus including event types and event texts obtained from grain storage pest texts after event extraction.

**LLM:**

We use ChatGLM [29] as the backbone large language model of our news summarization framework.

The effectiveness of the method proposed in this paper was verified by experimenting with two different tasks:

(1) The event pattern getter acquires and evolves the event pattern quality enhancement effect afterward.

(2) The experimental results of the framework on the niche domain dataset are compared with the PENS news headline generation dataset to demonstrate the generalization ability of the obtained news summaries.

**4.2 Results of News summary generator**

In this section, the quality of news summaries generated by NSG is evaluated, and the baseline selection of the TFIDF-based algorithm and TextRank-based algorithm for news summary generation. As well as a Gpt2-based news summary generation method [30], the main idea is to generate multiple summaries and select the most appropriate news summary from them, while we use LLM to generate summaries directly on the text.

Experimental evaluation metrics are used ROUGE, BLEU and overlap score, ROUGE is a text summarization evaluation metric based on n-gram and recall, which measures the quality of summarization by examining aspects such as word overlap, sentence-level similarity, and sequence-level similarity in summarization, ROUGE [31] metric system includes several evaluation metrics, such as ROUGE-1, ROUGE-2, and ROUGE-L, etc., which denote measuring one-word matching, measuring binary word matching, and recording the longest common subsequence, respectively, and are widely used in the evaluation and comparison of text summarization tasks. BLEU [32] is a precision-based similarity measure, and the common metrics are BLEU-1, BLEU-2, BLEU-3, and BLEU-4, Overlap detects the frequency of occurrence of repeated phrases between the generated news headlines and the reference headlines, and can measure to what extent the model is copying phrases directly from the text as summaries.

| Model | ROUGE | | | BLEU | | | | Overlap |
|---|---|---|---|---|---|---|---|---|
| | R-1 | R-2 | R-L | B-1 | B-2 | B-3 | B-4 | |
| TFIDF | 0.265 | 0.087 | 0.183 | 0.241 | 0.132 | 0.129 | 0.048 | 59 |
| TextRank | 0.354 | 0.143 | 0.282 | 0.312 | 0.188 | 0.174 | 0.082 | 52 |
| Gpt2 | 0.398 | 0.182 | 0.357 | 0.39 | 0.261 | 0.208 | 0.117 | 42 |
| GLM | 0.489 | 0.197 | 0.382 | 0.414 | 0.298 | 0.236 | 0.142 | 45 |
| GLM+NSG | 0.568 | 0.224 | 0.403 | 0.433 | 0.315 | 0.26 | 0.174 | 43 |

Table 1: Comparison of experimental results of GLM+NSG with baseline methods. We chose ROUGE, BLEU and Overlap as evaluation metrics, and the values of Overlap are percentages

We compare the news summary generation method of LLM plus NSG proposed in this paper with the above baseline most, the effect of LLM plus NSG for news summary generation is 0.431 and 0.417 higher than TFIDF and TextRank, respectively, which indicates that the method of TFIDF and PageRank weighting the adaptation of argumentative roles and then generating news summaries is better than directly using TFIDF and PageRank is better.

Also compared with Gpt2 and direct summary generation with a large model, the effects were 0.164 and 0.079 higher, respectively, demonstrating the effectiveness of extracting event patterns before population evolution.

### 4.3 Results in the Professional Field

The experimental effect on the grain storage domain dataset, the grain storage pest dataset is constructed for the purpose of constructing a grain storage pest matter mapping, and the event text paragraphs are short, and the data are in the form of grain storage pest related events and the corresponding matter, which is the abstract generalization of the corresponding event, which we use as the event title for the experiment.

| Model | ROUGE | | | BLEU | | | | Overlap |
|---|---|---|---|---|---|---|---|---|
| | R-1 | R-2 | R-L | B-1 | B-2 | B-3 | B-4 | |
| TFIDF | 0.195 | 0.094 | 0.122 | 0.062 | 0.105 | 0.116 | 0.052 | 63 |
| TextRank | 0.221 | 0.113 | 0.168 | 0.274 | 0.131 | 0.134 | 0.063 | 54 |
| Gpt2 | 0.256 | 0.127 | 0.219 | 0.241 | 0.163 | 0.152 | 0.072 | 38 |
| GLM | 0.314 | 0.133 | 0.273 | 0.284 | 0.179 | 0.163 | 0.088 | 40 |
| GLM+NSG | 0.378 | 0.172 | 0.305 | 0.312 | 0.202 | 0.193 | 0.112 | 42 |

Table 2: GLM+NSG results on stored grain pest dataset compared to baseline experiments

The experimental results on the corpus of stored grain pests we can see that the effect of the LLM plus NSG approach decreases on the PENS dataset, but the overall effect is still better than the baseline, which proves the generalization of the news summary extraction paradigm proposed in this paper. On the Overlap metric, since the methods based on TFIDF and TextRank are extractive methods, they are more dependent on the original data when the amount of data is small, so the obtained event summaries have a higher Overlap rate with the original text, while the methods based on GPT2 and LLM and the proposed method in this paper are generative, so the Overlap rate is reduced.

Experimental results show that the proposed method in this paper still achieves good results on the grain storage pest dataset, proving to be equally effective for summary generation tasks in niche domains.

### 4. Conclusion

In this paper, we propose a new paradigm for news summary generation and design a News Summary Generator (NSG). Experimental verify that the proposed approach in this paper is able to generate accurate and reliable news summaries on news summary generation tasks with certain generalizations and application to related tasks. At the same time, by introducing the event model into the news summary generation process of the LLM, the problem of its lack of interpretability can be solved to a certain extent, enabling us to understand and explain the internal operation of the model better, as well as the basis and logic of generating news summaries. This helps to improve the credibility and reliability of the LLM and provide more reliable and interpretable results for the news summary generation task.